\pdfoutput=1

\documentclass[11pt]{article}
\usepackage{graphicx}
\usepackage[]{ACL2023}
\usepackage{times}
\usepackage{latexsym}
\usepackage{lipsum}  
\usepackage[symbol]{footmisc}

\usepackage[T1]{fontenc}
\usepackage{booktabs}
\usepackage{multirow}
\usepackage{multicol}

\usepackage[utf8]{inputenc}

\usepackage{microtype}
\usepackage{inconsolata}

\usepackage[linewidth=1pt]{mdframed}

\title{\textit{Too Late to Train, Too Early To Use?} \\ A Study on Necessity and Viability of Low-Resource Bengali LLMs}

\author{
Tamzeed Mahfuz$^1$\thanks{~ Equal Contribution}, Satak Kumar Dey$^1$\footnotemark[1], Ruwad Naswan$^1$\footnotemark[1], Hasnaen Adil$^1$\\ \textbf{Khondker Salman Sayeed}$^2$, \textbf{Haz Sameen Shahgir}$^{3}$\thanks{~ Corresponding author: \texttt{hshah057@ucr.edu}}\\ [3pt]
Bangladesh University of Engineering and Technology$^1$, IQVIA$^2$, University of California Riverside$^3$
}

\begin{document}
\maketitle

\begin{abstract}
Each new generation of English-oriented Large Language Models (LLMs) exhibits enhanced cross-lingual transfer capabilities and significantly outperforms older LLMs on low-resource languages. This prompts the question: \textit{Is there a need for LLMs dedicated to a particular low-resource language?} We aim to explore this question for Bengali, a low-to-moderate resource Indo-Aryan language native to the Bengal region of South Asia.

We compare the performance of open-weight and closed-source LLMs such as LLaMA-3 and GPT-4 against fine-tuned encoder-decoder models across a diverse set of Bengali downstream tasks, including translation, summarization, paraphrasing, question-answering, and natural language inference. Our findings reveal that while LLMs generally excel in reasoning tasks, their performance in tasks requiring Bengali script generation is inconsistent. Key challenges include inefficient tokenization of Bengali script by existing LLMs, leading to increased computational costs and potential performance degradation. Additionally, we highlight biases in machine-translated datasets commonly used for Bengali NLP tasks. We conclude that there is a significant need for a Bengali-oriented LLM, but the field currently lacks the high-quality pretraining and instruction-tuning datasets necessary to develop a highly effective model.\footnote{~~The evaluation code is available at \url{https://github.com/satak100/BanglaBench}.}

\end{abstract}

\section{Introduction}
\begin{figure}[t]
    \centering
    \includegraphics[width=1\linewidth]{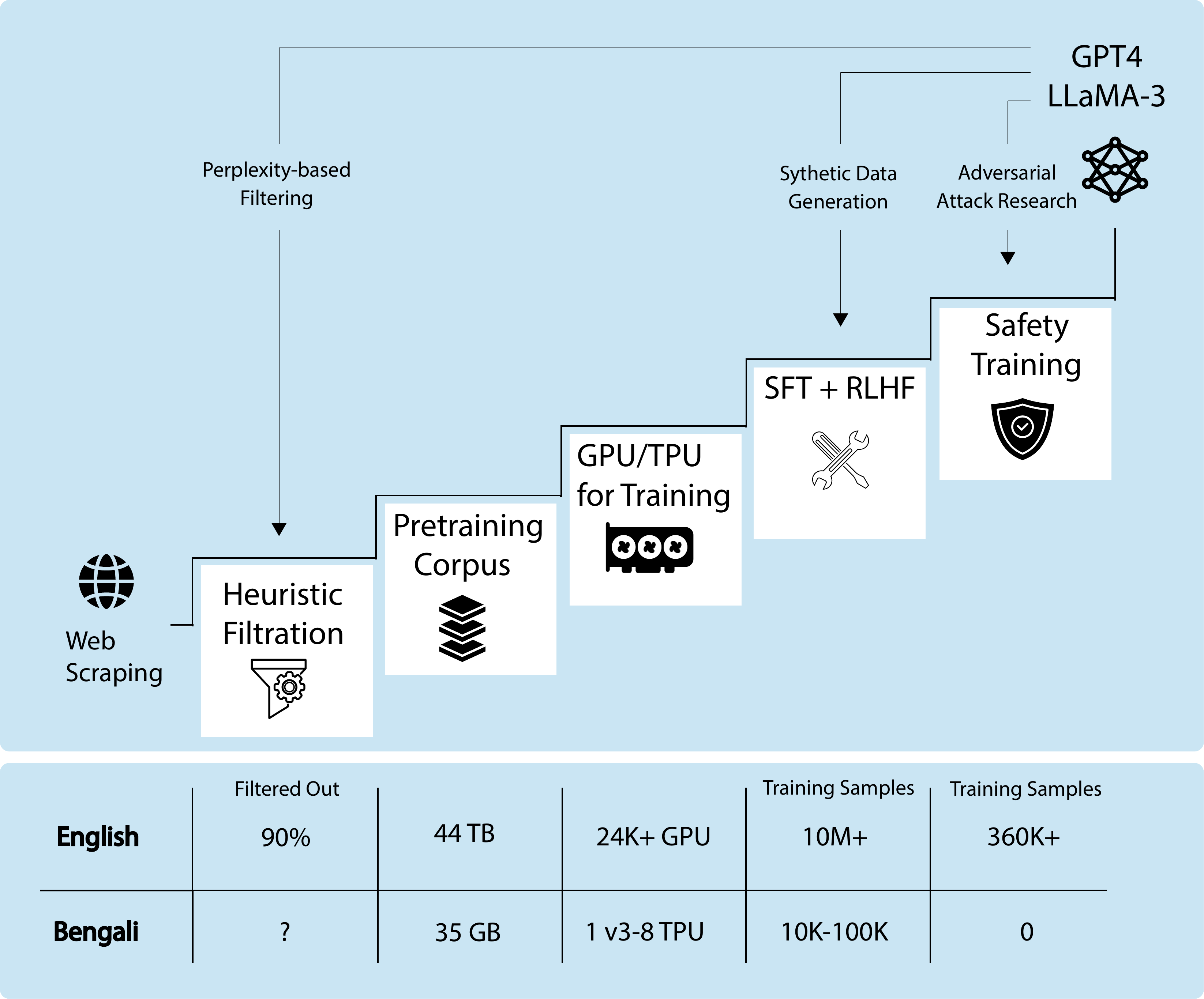}
    \caption{Large Language Model Training Pipeline and Resource Comparison between BanglaT5 \citep{banglat5} vs. LLaMA-3 \citep{llama3}. The first step towards capable Bengali LLMs is collecting a large pretraining corpus. However, the iterative nature of LLM development makes it unlikely that sufficient pretraining data and compute alone would enable Bengali LLMs to match the capabilities of their English-oriented counterparts.}
    \label{fig:open-fig}
\end{figure}

The release of GPT-3.5 \citep{fewshot} in late 2022 has kickstarted the current era of rapid progress in Large Language Models (LLMs). However, this progress is not merely a result of increased model scale, rather, it stems from a virtuous cycle of innovation, where lessons from each generation inform the development of the next. Techniques such as synthetic data generation \citep{eldan2023tinystories, gunasekar2023textbooks}, the integration of mathematical and coding tasks to enhance reasoning capabilities \citep{codeReasoning}, and research into adversarial attacks \citep{llmattack} for improved safety have all contributed to the ever-increasing capabilities of LLMs. As illustrated in Figure \ref{fig:open-fig}, developing state-of-the-art LLMs involves filtering vast amounts of web-scraped data, utilizing substantial computational resources, and implementing advanced techniques for alignment and safety.
\looseness=-1

However, this progress poses a dilemma for low-resource languages like Bengali. Despite being one of the most widely spoken languages, the size of Bengali pretraining and instruction-tuning data are minuscule compared to their English counterparts \citep{bangla_translation, banglabert}. To this date, BanglaT5 \citep{banglat5}, a 248 million parameter encoder-decoder T5 transformer \citep{t5}, remains the most capable Bengali Language Model. Furthermore, prematurely investing in training larger models might yield lackluster results due to the lack of high-quality Bengali data.
\looseness=-1

In this study, we aim to quantify the demand and viability of a Bengali-oriented LLM. To this end, we compile a representative benchmark of both Natural Language Understanding (NLU) and Natural Language Generation (NLG) downstream tasks for Bengali and evaluate a wide range of open-weights and closed-source models. Our key findings include:
\looseness=-1
\begin{enumerate}
    \item Compared to fine-tuned BanglaT5 or BanglaBERT, English-oriented LLMs excel in comprehension tasks (NLU) and perform inconsistently in Bengali generation (NLG).
    
    \item Using machine translation to translate English NLG datasets into Bengali biases the dataset towards specific writing styles and skews downstream metrics such as BLEU and ROUGE in favor of fine-tuned models regardless of generation quality.
    
    \item Bengali is over-tokenized by the BPE tokenizer used English LLM, with an average of $\sim0.85$ characters-per-token compared to $\sim4.5$ for English. Over-tokenization leads to $O(n^2)$ attention-based LLMs being highly inefficient in processing Bengali script.

    \item The outputs of English LLMs on Bengali Reward Modeling tasks do not correlate strongly with human judgment. As such, these LLMs have limited applicability in generating Bengali RLHF datasets.
    
\end{enumerate}
A comprehensive evaluation of state-of-the-art LLMs on 7 Bengali NLU and NLG tasks, revealing task-dependent performance variations.
An analysis of the inefficient tokenization of Bengali script by existing LLMs and its impact on model performance.
Insights into the challenges and potential strategies for developing Bengali-specific LLMs, balancing the need for language-specific models against the rapid progress in multilingual capabilities of existing LLMs.
\looseness=-1

\section{Preliminaries}
\subsection{Tasks and Datasets}
\label{sec:task_data}
We evaluate the latest LLMs on a wide range of Bengali downstream tasks as shown in Table \ref{tab:all_datasets}, covering both Natural Language Understanding (NLU) and Natural Language Generation (NLG). We elaborate on the differences between the Question-Answering datasets and the construction of the Reward Modeling dataset. 
\looseness=-1

\paragraph{Question-Answering:}
Among the Question-Answering (QA) datasets, Squad-bn \citep{banglabert} and BanglaRQA \cite{ekram2022banglarqa} are close-ended reading comprehension datasets, i.e. the LLM is given a context and a question, and must first determine whether the answer is present in the context and then extract the answer if it does. BEnQA is a close-ended, open-domain QA dataset where the LLM is asked a factual STEM-related question from the middle-school/high-school curriculum of Bangladesh.
\looseness=-1

\paragraph{Reward Modeling:} While combined fine-tuning downstream tasks \citep{flan} such as translation, summarization, and Question-Answering was the dominant post-pretraining paradigm from early Language Models such as T5 \citep{t5}, much of the impressible capabilities of Billion parameter-scale LLMs can be attributed to RLHF \citep{instructGPT, rlhf}, which improves the generalizability of LLMs even to unseen tasks. \citet{lee2023rlaif} has shown that feedback from other LLMs can substitute the need for human feedback in RLHF, in a method dubbed RLAIF. RLAIF can also be more robust than simple synthetic fine-tuning data generation \citep{abdin2024phi} which might overfit benchmarks \citep{overfit}. To test the capability of English LLMs to provide feedback on Bengali NLG, we created a new Reward Modeling task based on XLSum \citep{xlsum}. In this abstractive summarization dataset, we give the LLM a Bengali article and two summaries and ask it to pick the better one. We take the summary in the XLSum as the gold summary and the article's first sentence as the heuristically best summary. We instruct the LLM to prefer abstractive summaries over extractive ones. Refer to Appendix \ref{app:prompts} for the instruction template used. We randomly pick 300 samples from the test dataset due to cost considerations.
\looseness=-1

\begin{table*}[t]
    \centering
    
    \resizebox{1\textwidth}{!}{
    \begin{tabular}{c|llcccl}
        \toprule
        \textbf{Type} & \textbf{Task} & \textbf{Dataset} & \textbf{|Test|} & \begin{tabular}{@{}c@{}}\textbf{Data} \\ \textbf{Curation} \end{tabular} & \textbf{Metric} & \textbf{Best Model}\\
        \midrule
        \multirow{8}{*}{NLG} & Translation & \begin{tabular}{@{}l@{}}BanglaNMT \\  \citep{bangla_translation} \end{tabular} & 1000 & aligned & BLEU &\begin{tabular}{@{}l@{}}LLaMA-3-70B (B-E) \\  NLLB-3.3B (E-B) \end{tabular} \\ 
        \cmidrule{2-7}
         & \begin{tabular}{@{}l@{}}Monolingual \\ Summarization \end{tabular}   & \begin{tabular}{@{}l@{}}XLSum \\  \citep{xlsum} \end{tabular}   & 1012  & in-language  & ROUGE-2 & BanglaT5-248M-FT \\
        \cmidrule{2-7}
         &  \begin{tabular}{@{}l@{}}Crosslingual \\ Summarization \end{tabular} & \begin{tabular}{@{}l@{}}CrossSum \\ \citep{crosssum}  \end{tabular} & \begin{tabular}{@{}l@{}}161 (E-B) \\ 161 (B-E) \end{tabular} &  aligned & ROUGE-2 & \begin{tabular}{@{}l@{}} LLaMA-3-70B (E-B) \\ LLaMA-3-70B (B-E) \end{tabular} \\
        \cmidrule{2-7}
         & Paraphrase & \begin{tabular}{@{}l@{}}BanglaParaphrase \\ \citep{banglaparaphrase} \end{tabular}    & 	23332 & \begin{tabular}{@{}l@{}}machine  \\ translated \end{tabular} & ROUGE-2 & BanglaT5-248M-FT\\
        \cmidrule{1-7}
        \multirow{11}{*}{NLU} & QA (compr.) & \begin{tabular}{@{}l@{}}Squad-bn/BQA  \\ \citep{banglabert} \end{tabular}  & 2504 & \begin{tabular}{@{}l@{}}machine  \\ translated \end{tabular}   & F1/Match &  LLaMA-3-70B \\
        \cmidrule{2-7}
         & QA (compr.) & \begin{tabular}{@{}l@{}}BanglaRQA \\ \citep{ekram2022banglarqa} \end{tabular}  & 1493 & in-language & F1/Match &  LLaMA-3-8B-q4-FT \\
        \cmidrule{2-7}
         & QA (open-dom.) & \begin{tabular}{@{}l@{}}BEnQA \\ \citep{shafayat2024benqa} \end{tabular}   & 5161  & in-language & Acc. & GPT4\\
        \cmidrule{2-7}
         & Inference &   \begin{tabular}{@{}l@{}}XNLI-bn   \\\citep{banglabert} \end{tabular}  & 4895 & \begin{tabular}{@{}l@{}}machine  \\ translated \end{tabular} & Acc. & LLaMA-3-8B-q4-FT\\
         \cmidrule{2-7}
         & Reward Modeling &  \begin{tabular}{@{}l@{}}XLSum \\ (adapted subset) \end{tabular} & 300 & in-language & Acc. & LLaMA-3-70B \\
    
        \bottomrule
    \end{tabular}
    }
    \caption{Bengali datasets used in our experiments and the best model for each dataset. E-B stands for \textit{English-to-Bengali} generative tasks. FT stands for \textit{finetuned}.}
    \label{tab:all_datasets}
\end{table*}

\subsection{Models}
\label{sec:models}
Large Language Models can be categorized into open weights or closed-source models, based on whether individual users can download the model parameters or not. The current state-of-the-art LLM according to most benchmarks and user preference \citep{lmsys-chatbot-arena} is the closed-source GPT4o. The leading open-weights LLM in LLaMA-3-70B-Instruct \cite{llama3}, which ranks $9^{th}$ on the English-only \href{https://chat.lmsys.org/}{LMSYS Leaderboard} and $12^{th}$ overall. We note that open-weights models are not \textit{open-source} because most have proprietary licenses that restrict certain use cases such as commercial applications or synthetic data generation. 
\looseness=-1

\paragraph{Open-weights Models: } We test both the 8 and 70 billion variants of LLaMA-3 on all downstream tasks. We selectively include results from other open-weights LLMs such as Mistral-7B-0.3 \citep{jiang2023mistral}, Aya-23-8B \citep{aryabumi2024aya}, Qwen-2-72B \citep{qwen} for certain tasks. Aya-23 is a multilingual LLM family that was \textit{not} specifically trained for Bengali but was trained on related language families. Qwen-2 is a primarily English-Chinese LLM family with Bengali-specific training data augmentation \footnote{\url{https://qwenlm.github.io/blog/qwen2/}}. 
\looseness=-1

For translation, we test 3 variants of the translation-only NLLB Language Model \citep{costa2022no} on BanglaNMT \citep{bangla_translation}. We also test the performance of 8-bit quantized LLaMA-3-8B-Instruct to showcase what is possible on consumer-grade hardware. \textbf{All reported models are Chat- or Instruct-tuned unless specified otherwise.}
\looseness=-1

\paragraph{Closed-Source Models: } Due to high inference costs on Bengali text, we report GPT4o performance only on the Reward Modeling task. We report the performance of closed-source models such as GPT3.5, GPT4, and Gemini-1.5-Pro \cite{team2023gemini} if present in the literature. 
\looseness=-1

\subsection{Tokenization of Bengali Script}
\label{sec:tokenization}
Almost all LLMs use some variant of Byte-Pair Encoding (BPE) \citep{bpe}, an algorithm that iteratively combines the most common substrings into tokens. This naturally leads to under-represented scripts and notations being tokenized at higher granularities, leading to undertrained tokens, lower efficiency, and information density. 
\looseness=-1

We run pilot experiments on how LLMs tokenize Bengali text using the articles in XLSum \citep{xlsum}. We find that the average character-per-token value for Bengali using English LLMs is $\sim0.85$, which means that each token corresponds to \textit{less} than one Unicode Bengali character. In comparison, the character-per-token value for English is $\sim4.5$. The notable exception is BanglaT5 \citep{banglat5} which trained the tokenizer mainly on Bengali text and NLLB \citep{costa2022no}, which upsampled low-resource languages and downsampled high-resource ones when training the tokenizer. Detailed findings are presented in Appendix \ref{app:tokenization_comp}.
\looseness=-1

\citet{indic-extreme} notes the excessive tokenization of Bengali by BERT-based models. \citet{how_multilingual} highlight a novel link between excessive tokenization and the subpar performance of finetuning on languages that use a non-Latin script. They further highlight the existence of redundant tokens and show that removing them improves finetuning results.
\looseness=-1

\section{Experimental Setup}
We use \href{https://www.together.ai/}{Together AI API} for full-precision inference with open-weights LLMs. For NLLB and 8-bit quantized LLaMA-3-8B, we use \href{https://huggingface.co/}{Hugging Face} library on a single NVIDIA RTX A6000 machine. For Aya-23-8B, we use a $3\times$ NVIDIA RTX A6000 cluster. We use the \href{https://github.com/mjpost/sacrebleu}{sacreBLEU} library to calculate BLEU and the \href{https://github.com/csebuetnlp/xl-sum/tree/master/multilingual_rouge_scoring}{Multilingual-rouge-scoring} repository for ROUGE scores on Bengali text. For summarization tasks \citep{xlsum, crosssum}, we truncate the articles to 7000 tokens using the LLaMA-3 tokenizer due to the 8192 context window of LLaMA-3 models.
\looseness=-1

In tasks where frozen LLMs underperform, we minimally fine-tune LLaMA-3-8B-Instruct to probe the limitations of LLMs. We finetune LLaMA-3-8B-Instruct using 4-bit integer quantization, QLoRA \citep{dettmers2024qlora} using the \href{https://github.com/unslothai/unsloth}{Unsloth AI} library on a single NVIDIA RTX A6000. Task-wise hyperparameters are in Appendix \ref{app:finetuning}.
\looseness=-1

\section{Results}
In this section, we cover the results of our experiments on NLG and NLU tasks sequentially. The excellent performance of the NLLB-1.3B-Distilled \citep{costa2022no} on Bengali-to-English transition as highlighted in Section \ref{res:translation} and the lackluster performance of even GPT4o on Reward Modeling in Section \ref{res:pref_model} are particularly noteworthy since both are relevant to synthetic data generation and RLHF required to train LLMs.
\looseness=-1

\subsection{Translation}

Table \ref{tab:translation} shows that Google Translate significantly outperforms all LLMs and encoder-decoder transformers on both Bengali-to-English (B-E) and English-to-Bengali (E-B) translations. The large difference between Google Translate and other LLMs potentially points to data contamination. On the FLORES-101 \citep{goyal-etal-2022-flores} Bengali dev. test, NLLB-200 models were significantly better than Google Translate \citep{costa2022no}.
\looseness=-1

LLaMA-3-70B is the most capable B-E translator among the open-weights models, beating out the finetuned BanglaT5 \cite{banglat5}. This result disagrees with \citet{asai2023buffet} where small, finetuned encoder-decoder models outperformed LLaMA-2 on other tasks. Perhaps more impressively, the translation-specialized NLLB-3.3B \citep{costa2022no} is the best E-B translator, with even smaller NLLB variants outperforming much larger LLMs. As highlighted in Appendix Table \ref{tab:tokenization_comp}, The NLLB model family also boasts better tokenization support for Bengali, further improving inference speed and efficiency. Notably, the largest NLLB model is NLLB-54B-MoE which performs even better. See Appendix Table \ref{tab:nllb_only} for the comparison of NLLB variants on another English-Bengali dataset.
\looseness=-1

The consistent difference between E-B and B-E underlines how all translation systems find it harder to generate Bengali script (E-B) than to understand it (B-E).

\label{res:translation}
\begin{table}[!ht]
    \centering
    \begin{tabular}{l|c|c}
        \toprule
        \textbf{Model} & \textbf{B-E} & \textbf{E-B} \\
       \midrule
        BanglaT5-248M-FT                      & 31.30 & 17.40\\
        NLLB-600M-dis.                        & 29.52 & 17.56 \\
        NLLB-1.3B-dis.                        & 30.96 & 18.97 \\
        \vspace{0.5em}
        NLLB-3.3B                             & 30.97 & \textbf{19.73} \\

        Mistral-7B-v0.3                       & 14.91 & 3.67 \\
        LLaMA-3-8B-q8                         & 26.82 & 12.07\\ 
        LLaMA-3-8B                            & 28.48 & 12.82 \\
        LLaMA-3-70B                           & \textbf{33.55} & 18.92 \\
        \vspace{0.5em}
        Qwen-2-72B                            & 32.68 & 14.34 \\
        Google Translate     & 38.58$\dagger$ & 28.15$\dagger$   \\
        \bottomrule
    \end{tabular}
    \caption{Bengali-to-English (B-E) and English-to-Bengali (E-B) Translation performance of different models on BanglaNMT \citep{bangla_translation}. Reporting BLEU scores. $\dagger$ Google Translate API was used on June 21, 2024. The large BLEU score gap suggests data contamination in the Google Translate engine.}
    \label{tab:translation}
\end{table}

\subsection{Summarization}
In Table \ref{tab:all_sum}, we show that the finetuned BanglaT5 \citep{banglat5}, a 248M encoder-decoder performs better than even LLaMA-3-70B, a $320\times$ larger English LLM on Bengali-to-Bengali (B-B) summarization. B-B summarization requires both Bengali reading comprehension and generation. On the other hand, LLaMA-3-70B has $2\times$ higher ROUGE-2 score than BanglaT5 on B-E cross-lingual summarization and outperforms it on E-B summarization as well. Even the smaller 8B LLaMA-3 variant outperforms BanglaT5 on B-E CrossSum while Qwen-2-72B performs on par with the similarly sized LLaMA-3.
\looseness=-1
\begin{table}[!ht]
    \centering
    \resizebox{0.48\textwidth}{!}{
    \begin{tabular}{l|c|ccc}
        \toprule
        \textbf{Dataset} & \textbf{Model} & \textbf{B-B} & \textbf{B-E} & \textbf{E-B} \\
       \midrule
       \multirow{2}{*}{{XLSum}} & BanglaT5-FT    & \textbf{13.7}  & - &  - \\
        & Mistral-7B-v0.3                   &   6.40   & - &  - \\ 
        & LLaMA-3-8B                        &   7.36   & - &  - \\
        & LLaMA-3-70B                       &   8.66   & - &  - \\
        & Qwen-2-72B                        &   7.54   & - &  - \\
               \midrule
       \multirow{2}{*}{{CrossSum}}  & BanglaT5-FT & -  & 6.40 & 4.00\\
        & Mistral-7B-v0.3        & - & 5.61 & 3.21\\
        & LLaMA-3-8B             & - & 8.88 & 2.75 \\
        & LLaMA-3-70B            & - & \textbf{12.83} & \textbf{4.93} \\
        & Qwen-2-72B             & - & 12.54 & 4.91 \\
             \bottomrule
    \end{tabular}
    }
    \caption{ROUGE-2 scores of LLMs on XLSum \citep{xlsum} and CrossSum \citep{crosssum}. B-B denotes Bengali Article-to-Bengali summaries.}
    \label{tab:all_sum}
\end{table}

\subsection{Paraphrasing}
\label{subsec:paraph}

\begin{table}[!ht]
    \centering
    \begin{tabular}{l|c|c}
        \toprule
        \textbf{Dataset} & \textbf{Model} & \textbf{BLEU}\\
       \midrule
        & BanglaT5-FT    & \textbf{32.80} \\
        & LLaMA-3-8B-q8                  &   8.21   \\ 
       Bangla- & LLaMA-3-8B-q4-FT        &   26.99   \\ 
       Paraphrase & LLaMA-3-8B           &   9.13   \\
        & LLaMA-3-70B                    &   10.18   \\
        & Qwen-2-72B                     & 12.47 \\
             \bottomrule
    \end{tabular}
    \caption{Performance of different models on BanglaParaphrase \citep{banglaparaphrase}. }
    \label{tab:paraph}
\end{table}

Table \ref{tab:paraph} shows the finetuned BanglaT5 \citep{banglat5} outperforms all LLMs on Bengali paraphrase generation. As with B-B summarization, BanglaParaphrase \citep{banglaparaphrase} is also a Bengali-to-Bengali task. However, BanglaT5's BLEU metric is $3\times$ higher than even the LLaMA-3-70B. We manually inspected the reference paraphrase in the dataset and BanglaT5's and LLaMA-3-70B outputs. We discovered that the paraphrases generated by BanglaT5's outputs were more similar to the reference paraphrase in word choice, succinctness, and grammatical structure, while LLaMA-3-70B generated different but still perfectly valid paraphrases, with a slight tendency to generate longer phrases. Therefore, we suspect the high BLEU score of BanglaT5 to the fact that BanglaParaphrase was generated synthetically using translation and back-translation. Specifically, \citet{banglaparaphrase} used the translation model introduced by \citet{bangla_translation} to generate 5 paraphrases of each Bengali sentence in their corpus and filtered using LaBSE \citep{labse}. Both the translation pipeline and the choice of filtration likely introduce grammatical and word-choice biases into the dataset.
\looseness=-1

To investigate our suspicion, we run a small-scale fine-tuning experiment on LLaMA-3-8B-Instruct. We finetune LLaMa-3 using 4-bit quantization and QLoRA \cite{dettmers2024qlora} for only 1 epoch on the $420K$ training samples from BanglaParaphrase. \footnote{In contrast, BanglaT5 was fine-tuned for 10 epochs on $551K$ masking-augmented training samples in full-precision.} Despite using int-4 quantization and QLoRA, our fine-tuned LLaMA-3-8B-q4-FT significantly outperformed all non-finetuned LLMs including LLaMA-3-70B. Through manual inspection, we find that LLaMA-3-8B-q4-FT generates phrases similar to the reference paraphrase, with overlapping word choice and grammatical structure. Therefore, we surmise that the use of machine translation and LaBSE filtering has biased the reference summaries in Banglaphrase towards a certain linguistic style. As such, we advocate for human evaluation \citep{human_feedback} over automated metrics such as BLEU or ROUGE for synthetic NLG tasks. 
\looseness=-1

\subsection{Question-Answering}
\label{subsec:qa_results}
\begin{table}[!ht]
    \centering
    \resizebox{0.48\textwidth}{!}{
    \begin{tabular}{l|c|c|c}
        \toprule
        \textbf{Dataset} & \textbf{Model} & \textbf{F1} & \textbf{Exact} \\
       \midrule
       \multirow{2}{*}{Squad-Bn}& BanglaT5-FT  & 74.8  & 68.5  \\
        & Mistral-7B-v0.3                   & 54.9     & 49.8 \\
        & LLaMA-3-8B-q8                     &  75      & 68.5    \\
        & LLaMA-3-8B                        & 75.5     & 68.8 \\
        & LLaMA-3-70B                       & \textbf{81.9}     & \textbf{75.8} \\
        & Aya-23-8B                         & 36.8     & 29.4 \\
        \midrule
        \multirow{2}{*}{BanglaRQA}

        & BanglaBERT-FT                  &    63.2  & 47.6  \\ 
        & BanglaT5-FT                    &    78.1  & 62.4  \\ 
        & LLaMA-3-8B                     &    69.2  & 52.7  \\
        & LLaMA-3-8B-q4-FT               &    \textbf{80}  & \textbf{65.8}  \\ 
        & LLaMA-3-70B                    &    72.2  & 52.1  \\
        \midrule
        \multirow{2}{*}{BEnQA}& LLaMA-3-8B    &    -    &   45.7   \\
        & LLaMA-3-70B                       &      -    &   64.8   \\
        & GPT3.5$\dagger$                   &      -    &   37.2   \\
        & GPT4$\dagger$                    &      -    &   \textbf{75.1}   \\
        \bottomrule
    \end{tabular}
    }
    \caption{Bengali Question-Answering performance of LMs on Squad-bn \citep{banglabert}, BanglaRQA \citep{ekram2022banglarqa} and BEnQA \citep{shafayat2024benqa}. Reporting accuracy in the ``Exact" column for BEnQA. $\dagger$ Results from \citet{shafayat2024benqa}.}
    \label{tab:question-answering}
\end{table}

Out of the 3 QA datasets tested, Squad-Bn \cite{banglabert} and BanglaRQA \citep{ekram2022banglarqa} are reading comprehension tasks where a passage is provided and the models must answer with a single substring/span of the passage. Squad-bn and BanglaRQA have non-answerable questions, i.e. the answer is not in the passage. Furthermore, BanglaRQA contains questions where the answers are yes-no or multiple spans from the passage.
\looseness=-1

We instruct LLMs to determine if the answer exists in the context passage instead of answering directly from their parametric memory. For BanglaRQA, we instruct the LLM to determine the type of answer it should produce (yes-no, single-span, or multi-span) before writing the actual answer. See Appendix \ref{app:prompts} for the exact prompts used. Table \ref{tab:question-answering} shows that both LLaMA variants outperform the fine-tuned finetuned BanglaT5 on Squad-Bn. LLaMA-3-70B, in particular, shows convincing improvements in F1 (+7.1) and Exact Match (+7.3) metrics. However, in BanglaRQA, the fine-tuned BanglaT5 outperformed non-finetuned LLM by large margins. We manually inspected the LLaMA-3-70B's output and found it was prone to misclassifying yes-no and multiple-span questions as single-span questions. 
\looseness=-1

We fine-tuned LLaMA-3-8B-Instruct using QLoRA and 4-bit integer quantization for 3 epochs on the BanglaRQA train set. LLaMA-3-8B-q4-FT outperformed fine-tuned BanglaT5 by 1.9 units higher F1 and 3.4 percent higher Exact Matches.
\looseness=-1

BEnQA \citep{shafayat2024benqa} is an open-domain, multiple-choice QA dataset collected from the high school STEM curriculum of Bangladesh. Table \ref{tab:question-answering} shows that GPT4 leads LLaMA-3-70B by a significant margin (+$10.3$). Notably, the much smaller LLama-3-8B outperforms GPT3.5 by $8.5$ points, despite GPT3.5 and GPT4 using the same tokenizer. This suggests that the effect of over-tokenization \ref{sec:tokenization} is less pronounced on NLU tasks.  
\looseness=-1

\subsection{Natural Language Inference}

\begin{table}[!ht]
    \centering
    \begin{tabular}{l|c|c}
        \toprule
        \textbf{Dataset} & \textbf{Model} & \textbf{Acc.} \\
       \midrule
       \multirow{2}{*}{XNLI-bn} & BanglaBERT-FT    & \underline{82.8}\\
        & Mistral-7B-v0.3                   &    47.4  \\
        & LLaMA-3-8B-q8                     &    54.9  \\ 
        & LLaMA-3-8B-q4-FT                  &    \textbf{83.1} \\ 
        & LLaMA-3-8B                        &    57.3  \\
        & LLaMA-3-70B                       &    64.6  \\
        & Qwen-2-72B                        &    61  \\
        \midrule
        XNLI-bn $\dagger$ & GPT-3.5 Turbo &   92   \\
        \small(300 subset, 15-shot) & Gemini 1.5 Pro & 91.5\\
        \bottomrule
        
    \end{tabular}
    \caption{Bengali Natural Language Inference performance of LMs. $\dagger$ results from \citet{faria2024unraveling}.}
    \label{tab:inference}
\end{table}

Table \ref{tab:inference} shows a significant gap between finetuned and non-finetuned models on Natural Language Inference. Due to the large gap in performance between the finetuned BanglaBERT-111M \citep{banglabert} and LLaMA-3-70B, we minimally finetuned LLaMA-3-8B using parameter-efficient methods to probe the reason. Our finetuned LLaMA-3-8B-q4-FT even slightly outperforms BanglaBERT, showing that decoder-only LLMs can match encoder-only BERTs when finetuned. We additionally include results from \citet{faria2024unraveling}, where they find GPT-3.5 with 15-shot examples \citep{fewshot} significantly outperforms even finetuned models. We note that \citet{faria2024unraveling} only tested 300 random samples of XNLI-bn (out of 4895) due to the high cost associated with few-shot prompting.
\looseness=-1

\subsection{Reward Modeling}
\label{res:pref_model}

\begin{table}[!ht]
    \centering
    \begin{tabular}{c|c|c}
    \toprule
    \textbf{Dataset} & \textbf{Model} & \textbf{Acc.}\\
    \midrule
    \multirow{3}{*}{XLSum-en-300}   & LLaMA-3-8B & 58.67 \\
    & LLaMA-3-70B & \textbf{87.33}\\
    & GPT4o & \textbf{87.33}\\
    \midrule
    \multirow{3}{*}{XLSum-bn-300}   & LLaMA-3-8B & 53.67\\
    & LLaMA-3-70B & \textbf{67.67}\\
    & GPT4o & 63.33\\
    \midrule
    \multirow{2}{*}{\small{translated-XLSum-bn-300}}  & LLaMA-3-8B & 65.33\\
     & LLaMA-3-70B & \textbf{73.33}\\
    \bottomrule
    
    \end{tabular}
    \caption{Bengali Reward Modeling performance of LLMs. translated-XLSum-bn-300 denotes XLSum-bn-300 translated into English using NLLB-1.3-Distilled.}
    \label{tab:pref_model}
\end{table}
\vspace{-2mm}

As prefaced in Section \ref{sec:task_data}, we created a Reward Modeling task where we asked LLMs to choose the better summary of an article. See Appendix \ref{app:prompts} for the exact instruction used.

Table \ref{tab:pref_model} shows that LLaMA-3-8B largely fails to pick the correct summary, be it in English or Bengali. LLaMA-3-70B and GPT4o are evenly matched on the English dataset while LLaMA-3-8B performs close to random chance ($50\%$). We manually inspect LLaMA-3-8B's outputs and find that it prefers the verbatim nature of using the first sentence as the summary (Example LLaMA-3-8B output: \textit{``Summary 2 is better because it aligns closely to the article and does not include speculation or sensationalism."}). In Bengali articles, the performance of both LLaMA-3-70B and GPT4o degrade substantially. 
\looseness=-1

Since the output of reward models are usually language-agnostic, numeric, or binary values, we explore whether translating the Bengali article and summaries using an automated translator can recover the lost performance. Specifically, we translate the Bengali articles to English using NLLB-1.3B-Distilled \citep{costa2022no} and reattempt Reward Modeling on the translated dataset. This marginally recovers the accuracy of LLaMA-3-70B from $67.67\%$ to $73.33\%$. However, assuming humans have a $100\%$ accuracy on this task \footnote{A reasonable assumption since reference summaries were written by professional BBC contributors and the alternate summary is the article's first line.}, this wide gap between human and LLM preference bodes ill for using English LLMs as reward models for Bengali LLMs.
\looseness=-1

\section{Discussion}
In Section \ref{ssec:disc1}, we discuss potential issues with existing Bengali downstream tasks. In Sections \ref{ssec:disc2} and \ref{ssec:disc3}, we present key arguments for and against training a Bengali LLM in the short term.
\looseness=-1

\subsection{Pitfalls of Machine-Translated Datasets}
\label{ssec:disc1}
Table \ref{tab:all_datasets} shows that 3 out of 8 datasets we used were machine-translated. Machine translation is a cost-effective alternative to manual data annotation that requires much less human labor \citet{li2023cmmlu}. However, this risks translation errors being propagated through translated datasets, leading to second-order effects on LLM training and evaluation. Even if there are no errors, stylistic choices by automated translators can bias the dataset, something that is mitigated when there are multiple human annotators with different styles. We highlight such a case in Section \ref{subsec:paraph} on the BanglaParaphrase \citep{banglaparaphrase} dataset. 
\looseness=-1

\subsection{A Case for Training Bengali LLMs}
\label{ssec:disc2}

\paragraph{Better Generalization:} Our experiments show that English-only LLMs surpass fine-tuned BanglaT5 on NLU tasks while performing well in NLG datasets. Furthermore, \citet{asai2023buffet} shows that well-known emergent capabilities of monolingual LLMs such as Instruction Tuning \citep{instruction-tuning} and In-Context Learning \citep{fewshot} are less pronounced in other languages. 
\looseness=-1

\paragraph{Better Tokenization and Efficiency:} \citet{how_multilingual} shows that Bengali falls within the category of \textit{Stagnant Languages}, i.e. does not noticeably improve if finetuned. The authors suspect this stagnation against finetuning occurs in languages, including Bengali, that are tokenized excessively and therefore are information-sparse. Excessive tokenization is also harmful from a performance perspective due to the $\mathcal{O}(n^2)$ time complexity and  $\mathcal{O}(n)$ memory requirements of the standard attention mechanism in transformer-based LLMs.

\paragraph{Success of Chinese-oriented LLMs:} The rapid progress of Chinese and English-Chinese bilingual LLMs \citep{baichuan2023baichuan2, cai2024internlm2, deepseekv2, qwen} are particularly inspiring. Larger skews of Chinese LLMs such as Qwen-2-72B \citep{qwen} and DeepSeek-V2-236B-MoE \citep{deepseekv2} far outperform GPT4 \citep{achiam2023gpt} (90.1 vs. 70.95) on Chinese-MMLU \citep{li2023cmmlu}. Even smaller variants such as Baichuan2-13B \citep{baichuan2023baichuan2}, InternLM2-7B, and -20B \citep{cai2024internlm2} exhibit strong bilingual ICL capabilities.
\looseness=-1

The promise of a more capable and efficient Bengali Natural Language Generation, coupled with the proven success of Chinese LLMs are strong reasons to build a Bengali or English-Bengali LLMs. In fact, there have already been nascent attempts at such in the form of BanglaGPT \citep{banglagpt}, a GPT2-1.5B-based Bengali-only LLM.
\looseness=-1

\subsection{A Case Against Training Bengali LLMs}
\label{ssec:disc3}

\paragraph{Training Costs:} Although exact training costs have not been released, it is rumored that LLaMA-3-8B cost Meta around $5$ million USD on energy costs alone \citep{llama-cost-karpathy}. Meta built two custom 24K GPU superclusters\footnote{\url{https://ai.meta.com/blog/meta-llama-3/}} and trained LLaMA-3-8B $\times75$ longer than the Chinchilla \citep{hoffmann2022training} optimal point. Even more efficient architectures and training recipes such as JetMoE-8B \citep{shen2024jetmoe} required about $100$K USD to train and it performs significantly worse than LLaMA-3-8B.
\looseness=-1

\paragraph{Limited Bengali Data:} Beyond sheer training costs, the lack of high-quality Bengali datasets is another significant constraint. Currently, the largest Bengali pretraining corpus \citet{banglabert} is around 30GB while the largest open-source English corpus, FineWeb \citet{penedo2024fineweb} is 36.7 TB. Bengali also lacks the necessary RLHF datasets for instruction-tuning LLMs, a crucial step that aligns LLMs to human preferences and values.
\looseness=-1

The training of smaller LLMs such as LLaMA-3 \citep{abdin2024phi} or the Phi series \citep{abdin2024phi} is highly iterative and heavily dependent on being able to filter out low-quality data with older LLMs and generating high-quality synthetic (\textit{textbook quality}) data with larger LLMs such as GPT4 \citep{abdin2024phi}. Limited training data and the lack of preexisting Bengali LLMs create a negative feedback loop when attempting to train LLMs for Bengali.
\looseness=-1

\paragraph{Rapid Progress of Closed-source LLMs:} Any attempt to train a large-scale Bengali-oriented LLM may be premature due to the possibility of frontier AI labs increasing support for Bengali. For example, the latest model by OpenAI, GPT4o, reduced the token count of non-Latin scripts by as much as $4.4$ times compared to GPT4-Turbo \footnote{\url{https://openai.com/index/hello-gpt-4o/}}. Better Bengali support in frontier LLMs would significantly help synthetic data generation.
\looseness=-1

Building two-staged pipelines with state-of-the-art translation \citep{costa2022no} and English LLMs might be a better research direction in the short term while also being a significant stepping stone towards training LLMs for Bengali.
\looseness=-1

\subsection{Next Step towards Bengali LLMs: Compiling Pretraining Corpus}
Both BanglaT5 \citep{banglat5} and BanglaBERT \citep{banglabert} were trained on a new pretraining corpus Bangla2B+, collected by the authors via web-crawl on 110 Bengali websites. Alternative data sources include OSCAR \citep{oscar} and CCNet \citep{wenzek2020ccnet} but they reportedly contain noise and offensive text that is infeasible to filter out.

Bangla2B+ totals about 35 GB of data, a far cry from English pretraining corpora of 44 TB. Expanding the selection of scraped websites, transcribing Bengali media content, digitizing printed media, and translating English corpora using automated translation pipelines are possible methods to increase the Bengali pretraining corpus. As we have highlighted in \ref{subsec:paraph}, naive translation induces bias. We advocate for using multiple translation models in tandem, advanced prompting, and rigorous post-processing to ensure not just accuracy, but also linguistic naturalness and consistency.

\section{Other Related Works}
Alongside models pretrained exclusively on Bengali script such as BanglaT5 \citep{banglabert} and BanglaBERT \citep{banglat5}, there exists multilingual models trained on related Indic languages including Bengali. These include encoder-only transformers such as MuRILBERT \citep{khanuja2021muril}, IndicBERT \citep{indic-extreme}, and encoder-decoder transformers such as IndicBART \citep{dabre2021indicbart}.
\looseness=-1

Besides the datasets in Table \ref{tab:all_datasets}, other Bengali downstream tasks include grammatical error detection and correction, sentiment analysis, and transliteration. \citet{error-correction} introduced a new Bengali error detection dataset and correction and found that BanglaBERT \citep{banglabert} excels at detection while BanglaT5 \citep{banglat5} excels at correction. However, \citet{shahgir-t5-detection} finds that BanglaT5 can match BanglaBERT on detection, albeit on a different dataset \citep{error-detection}. \citet{sa-noise-reduction} finds that BenglaBERT outperforms MuRIL \citep{khanuja2021muril} on both noisy and noise-reduced Bengali sentiment analysis \citep{sa-noise}. \citet{roark-etal-2020-processing} introduces a Bengali transliteration dataset and finds that a transformer \citep{new-transformer} outperforms LSTMs at the task.
\looseness=-1

\citet{asai2023buffet} compares downstream task performance in multiple languages including Bengali. The authors find that in-context learning with LLMs such as BLOOMZ-7B, BLOOM-176B \citep{workshop2022bloom} and GPT-3.5 underperform compared against fine-tuned mT5 \citep{muennighoff2022crosslingual} baselines. Similarly, a concurrent work \citet{banglabench0} finds that fine-tuned BanglaT5 and BanglaBERT outperforms GPT-3.5, LLaMA-2-7B \citep{llama2} and Claude 2 \footnote{\url{https://www.anthropic.com/news/claude-2}}. In contrast, we test more recent and capable LLMs including LLaMA-3 \citep{llama3}, GPT4 \citet{achiam2023gpt} and find that LLMs outperform fine-tuned models in multiple Bengali benchmarks.
\looseness=-1

\section{Conclusion}
Our experiments reveal a mixed landscape; while LLMs generally outperform fine-tuned T5 baselines on Bengali NLU tasks, their performance on Bengali NLG tasks, particularly those requiring Bengali script generation, leaves room for improvement. Key findings include the inefficient tokenization of Bengali script by existing LLMs, task-dependent performance variations, and potential biases in machine-translated datasets. The study also highlights the significant costs and data requirements for training Bengali-specific LLMs, balanced against the rapid progress in the multilingual capabilities of existing models. In the short term, leveraging state-of-the-art translation models with powerful English LLMs may offer a pragmatic approach to improving Bengali language technologies. Future research should focus on developing more efficient tokenization methods for non-Latin scripts, creating high-quality Bengali datasets, and utilizing cross-lingual transfer.
\section{Limitations}

\paragraph{Lack of Human Evaluation}: While we identified the need for human evaluation in tasks like paraphrasing, we did not conduct human evaluations ourselves due to resource constraints. This limits our ability to fully assess the quality of model outputs, especially for generation tasks.

\paragraph{Tokenization Analysis}: Although we identified inefficiencies in Bengali script tokenization, a more in-depth analysis of its impact on model performance across different tasks and model sizes could provide further insights.
    
\paragraph{Fine-tuning Experiments}: The evaluation of larger models was limited by available computational resources. Our fine-tuning experiments were limited in scope and primarily focused on LLaMA-3-8B. A more comprehensive exploration of fine-tuning across different model architectures and sizes could yield additional insights.
    
\paragraph{Temporal Limitations}: Given the rapid pace of development in the field of LLMs, some of our findings may become outdated as new models and techniques are introduced.

\bibliography{custom}
\bibliographystyle{acl_natbib}

\newpage
\onecolumn

\appendix

\section{Finetuning Hyperparameters}
\label{app:finetuning}
We finetune LLaMA-3-8B-Instruct using QLoRA ($r=16$, $\alpha=16$) and 4-bit integer quantization with learning rate $5\times10^{-5}$, warmup-ratio $0.05$ and linear-rate scheduling for all tasks. We use a single Nvidia RTX A6000 for all experiments. 

On BanglaParaphrase \cite{banglaparaphrase}, we train for 1 epoch with batch size 32 and gradient accumulation every 4 batches. 

For BanglaRQA \citep{ekram2022banglarqa}, we train for 3 epochs with batch size 4 and gradient accumulation every 8 batches. We filtered out training and validation samples with context smaller than 500 and longer than 3900 characters for efficient training. 

For XNLI\_bn \citep{banglabert}, we train for 1 epoch with batch size 32 and gradient accumulation every 4 batches. We filtered out training samples where the combined length of the two sentences was less than 50 and longer than 350 characters for efficient training. 

We did not extensively tune hyperparameters for any fine-tuning experiments.
\section{Prompts}
\label{app:prompts}

\paragraph{Translation}

\begin{mdframed}[]
\begin{verbatim}
SYSTEM:
You are a state-of-the-art AI assistant that translates sentences from {Language A} 
to {Language B}. The user provides you with a {Language A} sentence, and your task is 
to translate it into {Language B}. Just return the translation without any preamble, 
quotations, or explanations.

USER: {Language A sentence}
\end{verbatim}
\end{mdframed}

\vspace{4em}

\paragraph{Paraphrase Generation}
\begin{mdframed}
\begin{verbatim}
SYSTEM:
You are a state-of-the-art AI assistant that generates Bengali paraphrases. The user 
provides you with a Bengali sentence, and your task is to generate a Bengali 
paraphrase of it. Just return the paraphrase without any preamble, quotations, 
or explanations.

USER: {Bengali sentence}
\end{verbatim}    
\end{mdframed}

\vspace{4em}
\paragraph{Summarization}

\begin{mdframed}
\begin{verbatim}
SYSTEM:
Please write a one-sentence {Language A} summary/TL;DR of the given {Language B} 
article. The summary must be in {Language A} and not be longer than a sentence. 
Just return the summary without any preamble, quotations, or explanations.

USER: {Language B Passage}
\end{verbatim}    
\end{mdframed}

\newpage
\paragraph{Natural Language Inference}
\begin{mdframed}
\begin{verbatim}
SYSTEM:
You will be given two sentences. Please determine whether the first sentence
entails, contradicts, or is neutral to the second. Pay close attention to each 
word as you analyze the relation between the two sentences. Respond in the format: 
Thought: {thought on if the first second entails, contradicts, or is neutral 
to the second sentence}\n\nVerdict: {any one of <entailment>, 
<contradiction> or <neutral> tags}

USER:
Sentence 1: {}\n\nSentence 2: {}
\end{verbatim}
\end{mdframed}
\vspace{4em}

\paragraph{BQA/Squad-bn}
\begin{mdframed}
\begin{verbatim}
SYSTEM: 
Is the to the question in the context? ('YES'/'NO'). What is the answer? (A substring
of the context/'<NOT_IN_CONTEXT>'). Return as a tuple (eg. ('YES', answer_substring
) or ('NO','<NOT_IN_CONTEXT>') without any preamble or explanations.

USER: Context: {context} \n\n Question: {question}
\end{verbatim}    
\end{mdframed}
\vspace{4em}
\paragraph{BanglaRQA}
\begin{mdframed}
\begin{verbatim}
SYSTEM:
The user will provide a context and a question, both in Bengali.
Read the context and the question carefully.

Respond with a JSON object with the following keys:

"answerable" (boolean, Is the question answerable from the context?)
"question_type" (yes-no / single-span / multiple-span)
"answer" ('Yes' or 'No' for yes-no questions)/substring of the context for single-
span/list of substrings of the multiple-span/'<NOT_IN_CONTEXT>')

USER:
Context: {}\n\nQuestion: {}
\end{verbatim}
\end{mdframed}

We used the Bengali words for 'Yes' and 'No' when specifying the \textit{answer} key in the above prompt.

\newpage
\paragraph{Reward Modeling}
\begin{mdframed}
\begin{verbatim}
UESR:
Here is a news article: 
<article>
{article}
</article>

Here is one person's summary of the article:
<summary1>
{summary1}
</summary1>

And here is a second person's summary of the same article:
<summary2>
{summary2}
</summary2>

Please read the article and both summaries carefully. Then, in <thoughts> tags, 
analyze the strengths and weaknesses of the two summaries, focusing on the following 
criteria:

1) Faithfulness - does the summary accurately reflect the key points of the article 
without adding extraneous or false information? 
2) Coherence - is the summary well-structured and easy to understand?
3) Concision - does the summary capture the essence of the article efficiently, 
without unnecessary detail or repetition?
4) Abstraction - does the summary rephrase the article content in novel ways, or does 
it just extract verbatim snippets? 

Favor summaries that demonstrate abstraction and rephrase content in their own words 
over ones that just extract verbatim snippets.

After you've thought it through, provide your final verdict on which summary 
is better inside <verdict> </verdict> tags, using either a <first> or <second> tag 
to indicate your choice. You must pick one or the other, you cannot hedge or say they 
are equal. The summary that does a better job meeting the above criteria, especially 
abstraction, should be selected as the better one.
\end{verbatim}
\end{mdframed}

\newpage
\paragraph{BEnQA}

\begin{mdframed}
\begin{verbatim}
SYSTEM:
You are given a multiple choice question and their options in English/Bengali and
your job is to correctly answer the question. First reason step by step in English/
Bengali and only then give me the final answer as "a", "b", "c" or "d".

Keep these in mind:
1. Only include the letter a, b, c, d as your final answer. Do not include the option 
text.
2. Every question will have an answer in the given options. So, DO NOT say that none 
of the answers are correct.
3. ONLY ONE of the given options will have the answer. So DO NOT provide multiple 
options as answers.
4. The questions contain enough information to solve the problem, so DO NOT say that 
you need additional information.
5. Answer in the format:
\n'Let's think step by step.\n{reasoning}\n\nAnswer:{A/B/C/D}'

USER: 
Question:
{Bengali question}

Options:
{Bengali options}
\end{verbatim}
\end{mdframed}

\newpage
\section{Tokenization of Bengali Script by English-oriented Language Models}
\label{app:tokenization_comp}

\begin{table}[h]
    \centering
    \begin{tabular}{lcccc}
        \toprule
        \textbf{Tokenizer} & \(|Context|\) & \(|Vocab|\) & \textbf{English} & \textbf{Bengali} \\
        \midrule
        BanglaT5 & 512 & 32K & 3.05 & 5.09 \\
        NLLB & 1024 & 256K & 4.25 & 3.35 \\
        AYA-23 & 8192 & 255K & 4.75 & 0.87 \\
        LLaMA-3 & 8196 & 128K & 4.77 & 0.83 \\
        Mistral & 32768 & 32K & 4.31 & 0.90 \\
        Qwen2 & 131072 & 152K & 4.69 & 0.94 \\
        \bottomrule
    \end{tabular}
    \caption{Average character per token values of different tokenizers on 11535 English and 8012 Bengali BBC articles from XLSum \citep{xlsum}. }
    \label{tab:tokenization_comp}
\end{table}

\begin{figure}[h]
    \centering
    \includegraphics[width=1\linewidth]{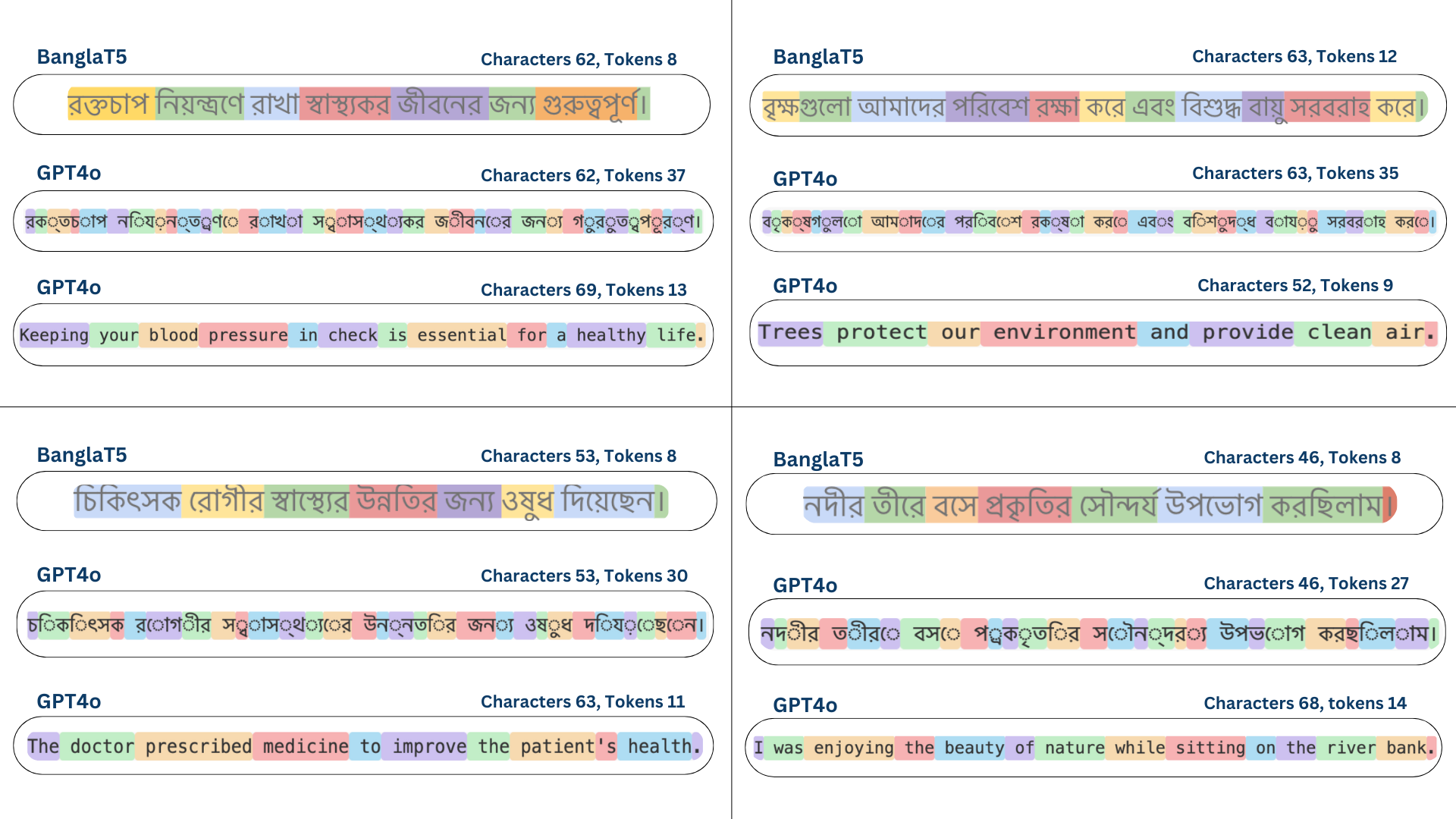}
    \caption{Qualitative Examples of Inefficient Bengali Script Tokenization.}
\end{figure}

\newpage
\section{Additional Results}

\subsection{BEnQA}

\begin{table}[!ht]
\centering
\begin{tabular}{ l | c | c c |c c}
\toprule
\textbf{Subject} & \textbf{Total}& \multicolumn{2}{c|}{\textbf{LLaMA-3}}  &\textbf{GPT3.5}  &\textbf{GPT4}\\

 & & 8B & 70B  & & \\
\midrule
8th-Math         & 209  &  0.584  & 0.722 &  0.486  &   0.808  \\
8th-Science      & 228  &  0.465  & 0.640 &  0.356  &   0.721  \\
10th-Biology     & 351  &  0.499  & 0.638 &  0.351  &   0.775  \\
10th-Chemistry   & 389  &  0.494  & 0.658 &  0.404  &   0.741  \\
10th-Math        & 380  &  0.453  & 0.700 &  0.407  &   0.775  \\
10th-Math-II     & 393  &  0.478  & 0.695 &  0.383  &   0.781  \\
10th-Physics     & 319  &  0.47   & 0.639 &  0.36   &   0.75  \\
12th-Biology-I   & 310  &  0.445  & 0.603 &  0.346  &   0.721  \\
12th-Biology-II  & 328  &  0.415  & 0.598 &  0.315  &   0.712  \\
12th-Chemistry-I & 367  &  0.469  & 0.638 &  0.314  &   0.775  \\
12th-Chemistry-II & 389 &  0.393  & 0.640 &  0.355  &   0.751  \\
12th-Math-I      & 396  &  0.467  & 0.684 &  0.431  &   0.756  \\
12th-Math-II     & 391  &  0.394  & 0.542 &  0.391  &   0.662  \\
12th-Physics-I   & 304  &  0.457  & 0.664 &  0.375  &   0.774  \\
12th-Physics-II  & 333  &  0.429  & 0.670 &  0.319  &   0.775  \\
12th-Chemistry-I & 367  &  0.469  & 0.638 &  0.314  &   0.775  \\
\midrule
\textbf{Total/Avg} & \textbf{5087} & \textbf{0.457} & \textbf{0.648} & \textbf{0.372} & \textbf{0.751} \\
\bottomrule
\end{tabular}
\caption{Subject-wise Accuracy in English.}
\label{table:BEnQA_accuracy}
\end{table}

\subsection{NLLB}
\begin{table}[!ht]
\centering
\begin{tabular}{l l c c c}
\toprule
\textbf{Model} & \textbf{Arch.} & \textbf{|Parameters|} & \textbf{E-B} & \textbf{B-E} \\
\midrule
NLLB-200 & MoE & 54.5B & 50.0 & 62.2 \\
NLLB-200 & Dense & 3.3B & 48.7 & 61.1 \\
NLLB-200 & Dense & 1.3B & 47.3 & 59.8 \\
NLLB-200-Distilled & Dense & 1.3B & 47.8 & 60.1 \\
NLLB-200-Distilled & Dense & 600M & 46.2 & 57.9 \\
\bottomrule
\end{tabular}
\caption{Translation Metric of the current state-of-the-art NLLB model family on the NLLB dataset \citep{costa2022no}. Reporting \href{https://github.com/m-popovic/chrF}{chrF++} scores. }
\label{tab:nllb_only}
\end{table}

\end{document}